\def\arxiv{}

\ifdefined\arxiv

    \documentclass[a4paper,10pt]{article} 
    
    \usepackage{natbib}
    \usepackage[utf8]{inputenc}
    \usepackage[T1]{fontenc}
    \usepackage{geometry}
    \usepackage{lipsum} 
    \usepackage{parskip} 
    \usepackage{xcolor}
    
    \setcitestyle{authoryear,round,citesep={;},aysep={,},yysep={;}}
    
    \usepackage{etoolbox}
    \ifundef{\abstract}{}{\patchcmd{\abstract}%
        {\quotation}{\quotation\noindent\ignorespaces}{}{}}

    \usepackage[backref=page,hyperfootnotes=false]{hyperref}
    \hypersetup{
      colorlinks,
      urlcolor  = Blue9,
      linkcolor = Blue9,
      citecolor = Blue9,
    }
        
\else

    \ifdefined\neurips
        \documentclass{article}

        \usepackage[final]{neurips_2023}
        
        \usepackage[utf8]{inputenc} 
        \usepackage[T1]{fontenc}    
        \usepackage{hyperref}       
        \usepackage{url}            
        \usepackage{booktabs}       
        \usepackage{amsfonts}       
        \usepackage{nicefrac}       
        \usepackage{microtype}      
        \usepackage{xcolor}         
        \usepackage{booktabs}
        \usepackage{multirow}
        \usepackage{wrapfig}
        \usepackage{lipsum}
        \usepackage{caption}
        \usepackage{caption} 
        \usepackage{graphicx} 
        \usepackage{adjustbox}
        \usepackage{array, makecell} %

    \else
    
        \documentclass{article} 
        \usepackage{iclr2024_conference}
        
        \usepackage{titlesec}
        
        \usepackage[backref=page]{hyperref}
        \hypersetup{
          colorlinks,
          urlcolor  = Blue9,
          linkcolor = Blue9,
          citecolor = Blue9,
        }
            
        \titlespacing\section{0pt}{4pt plus 2pt minus 2pt}{2pt plus 2pt minus 2pt}
            
    \fi
\fi

\usepackage{times}

\usepackage{amsmath,amsfonts,bm}

\def\eqref#1{equation~\ref{#1}}

\def\1{\bm{1}}

\DeclareMathAlphabet{\mathsfit}{\encodingdefault}{\sfdefault}{m}{sl}
\SetMathAlphabet{\mathsfit}{bold}{\encodingdefault}{\sfdefault}{bx}{n}

\title{Fine-tuning Language Models for Factuality}

\usepackage{url}
\usepackage{booktabs}
\usepackage{multirow}
\usepackage{wrapfig}
\usepackage{lipsum}
\usepackage{caption}
\usepackage{subcaption}
\usepackage{graphicx} 
\usepackage{adjustbox}
\usepackage{array, makecell} %

\usepackage{tabularx}
    \newcolumntype{L}{>{\raggedright\arraybackslash}X}

\definecolor{Blue9}{rgb}{0.098,0.3,0.9}

\author{\textbf{Katherine Tian\textsuperscript{\textasteriskcentered\textdagger}}, \textbf{Eric Mitchell\textsuperscript{\textasteriskcentered\textdagger}}, \textbf{Huaxiu Yao\textsuperscript{\textdagger\S}},\\\textbf{Christopher D. Manning\textsuperscript{\textdagger}}, \textbf{Chelsea Finn\textsuperscript{\textdagger}}\\\textsuperscript{\textdagger}Stanford University\hspace{2mm}\textsuperscript{\S}UNC Chapel Hill\\\texttt{\{kattian,eric.mitchell\}@cs.stanford.edu}}

\date{}

\begin{document}

\maketitle

\begin{abstract}
The fluency and creativity of large pre-trained language models (LLMs) have led to their widespread use, sometimes even as a replacement for traditional search engines. Yet language models are prone to making convincing but factually inaccurate claims, often referred to as `hallucinations.' These errors can inadvertently spread misinformation or harmfully perpetuate misconceptions. 
Further, manual fact-checking of model responses is a time-consuming process, making human factuality labels expensive to acquire. 
In this work, we fine-tune language models to be more factual, without human labeling and targeting more open-ended generation settings than past work. We leverage two key recent innovations in NLP to do so. First, several recent works have proposed methods for judging the factuality of open-ended text by measuring consistency with an external knowledge base or simply a large model's confidence scores. Second, the direct preference optimization algorithm enables straightforward fine-tuning of language models on objectives other than supervised imitation, using a preference ranking over possible model responses. We show that learning from automatically generated factuality preference rankings, generated either through existing retrieval systems or our novel retrieval-free approach, significantly improves the factuality (percent of generated claims that are correct) of Llama-2 on held-out topics compared with RLHF or decoding strategies targeted at factuality. At 7B scale, \textbf{compared to Llama-2-chat, we observe 58\% and 40\% reduction in factual error rate} when generating biographies and answering medical questions, respectively.

\end{abstract}

\section{Introduction}
\let\oldthefootnote\thefootnote
\renewcommand{\thefootnote}{\textasteriskcentered}
\footnotetext{Equal contribution.}\setcounter{footnote}{0}\let\thefootnote\oldthefootnote Recent developments in training large language models (LLMs), particularly methods that learn from rankings over responses such as reinforcement learning from human feedback (RLHF) \citep{christiano2017deep,ziegler2020finetuning,ouyang2022training}, have enabled the development of powerful, engaging dialogue agents. State-of-the-art LLMs are pre-trained on a vast amount of knowledge in large datasets \citep{touvron2023llama, touvron2023llama2} and further fine-tuned to apply this knowledge to follow diverse instructions or complete more specific tasks \citep{chung2022scaling,Chen2021EvaluatingLL}. However, despite these large language models' exposure to diverse datasets, they are prone to confidently generating incorrect claims. One recent study shows that GPT-3.5 (ChatGPT) produces false citations more often than not when asked to provide the authors of a given study \citep{agrawal2023language}. Nonetheless, other research has demonstrated that in simple question-answering settings, large language models \textit{do} exhibit systematic markers of uncertainty that indicate their factually unreliable statements \citep{kadavath2022language,tian2023just}. These results suggest that language models internally represent the limits of their knowledge, leading us to ask: \textit{Can language models be fine-tuned to leverage this internal awareness, to avoid making untrue statements in the first place?}

A key source of difficulty in training factual models comes in specifying an objective that adequately captures factuality. As an example, maximum likelihood, the most common objective for pre-training language models, does not always encourage factual predictions. Consider the question ``Where was Yo-Yo Ma born?'' A model that continues by near-deterministically producing the text ``idk, probably Paris?'' is nearly always correct, but receives extremely high loss if the pre-training data contains any other response to the question. On the other hand, a model that hedges probability mass over many possible phrasings and many possible locations (including incorrect ones, like Antarctica) will likely receive much lower loss, as any response observed in the training data will be assigned at least \textit{some} non-trivial probability. Because the pre-training objective may reward `smearing' probability mass over many possible responses, language models may generate incorrect statements if they underfit the training data or if asked questions that require knowledge not contained in the pre-training data.

In principle, reinforcement learning-based objectives can avoid the failures of existing pre-training objectives through the appropriate choice of a reward function that penalizes factually incorrect statements. However, accurately computing such a reward function can be expensive.
Obtaining human labels of factuality is time-consuming and costly; 
\citet{min2023factscore} report that professional fact-checkers took approximately 9 minutes to fact-check a single model-generated biography of a well-known individual; it cost about \$2,000 to annotate 505 biographies.

\begin{figure}
    \centering
    \includegraphics[width=\textwidth]{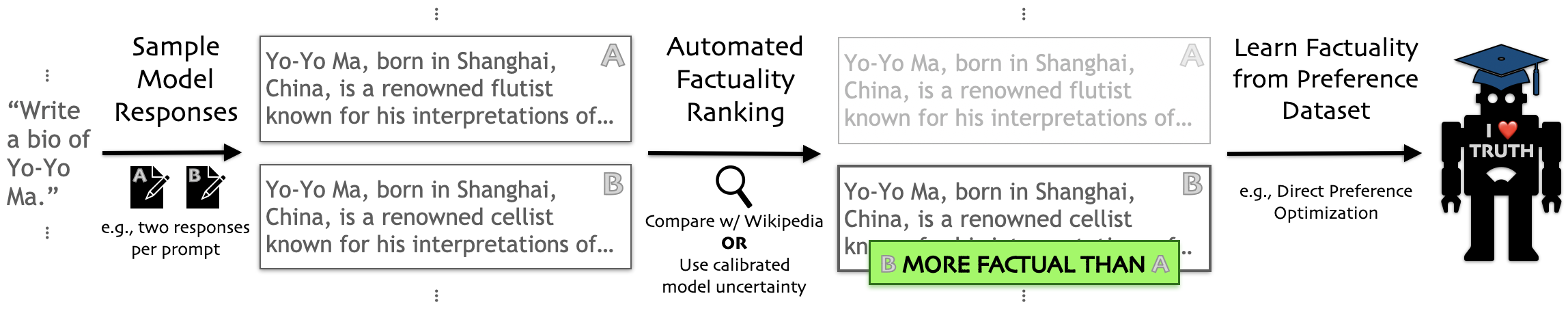}
    \caption{Our approach aims to improve the factuality of language models, specifically focusing on long-form generation (e.g. writing a biography). We develop two different approaches for estimating factuality of a passage (center), each of which allows us to generate a preference dataset (right). We then fine-tune the language model to optimize these factuality preferences (far right).
      }
    \label{fig:fig1}
\end{figure}

In light of these challenges, we leverage recent advances in estimating truthfulness \textbf{without human intervention}: a) \textit{reference-based} automated fact-checking methods that evaluate the extent to which an external knowledge base supports the claims in a piece of text \citep{min2023factscore,chern2023factool} and b) \textit{reference-free} truthfulness evaluations that use a model's own confidence as a proxy for truthfulness, inspired by \citet{kuhn2023semantic}. Using these truthfulness measures and a dataset of unlabeled prompts (e.g., ``Write a biography of Yo-Yo Ma.''), we sample pairs of completions from a pre-trained model and annotate them with a preference label denoting which has a lower rate of factual errors. Using the recently proposed Direct Preference Optimization \citep{rafailov2023direct} algorithm, we can stably and efficiently learn from such data. Ultimately, this pipeline enables us to fine-tune off-the-shelf language models to produce factual errors less often (with or without a reference knowledge base). See Figure~\ref{fig:fig1} for an overview of our factuality tuning pipeline.

Our primary contribution is a straightforward approach to optimizing language models for factuality in long-form text generation without human annotation. We validate this approach on two benchmark datasets for evaluating factuality, targeted at generating biographies of popular figures and answering open-ended questions about medical conditions. We find that fine-tuning for factuality outperforms conventional RLHF and produces complementary benefits to LLM decoding strategies that aim to increase factuality. Further, we find qualitative differences in the result of learning from preference pairs scored with reference-based and reference-free truthfulness estimation. Overall, we find that learning factuality from automatically constructed preference pairs is a cost-effective way to increase model factuality without human intervention, reducing the error rate for claims generated by Llama models by over 50\% for biographies and 20--30\% for medical questions.

\newcommand{\piref}{\pi_\text{ref}}

\section{Preliminaries}

Our approach to fine-tuning directly for improved factuality uses the framework of reinforcement learning from preferences over candidate actions or responses. In this section, we provide an overview of reinforcement learning in the context of language models, as well as the specific algorithm we use for preference-based RL, direct preference optimization \citep{rafailov2023direct}.

\textbf{Fine-tuning language models with reinforcement learning.} Reinforcement learning (RL) has proven to be an effective approach to fine-tuning language models to extract complex, useful behaviors from their pre-trained weights. In the context of RL, a language model policy $\pi_\theta$ (typically an autoregressive Transformer) produces a conditional distribution $\pi_\theta(y\mid x)$ over responses $y$ given an input query $x$ (both $x$ and $y$ are text sequences). The goal of reinforcement learning is to maximize the average reward of outputs generated by the policy, where a reward function $r(x,y)$ assigns a scalar score to an input-output pair that determines its desirability. However, past works have observed that fine-tuning language models with an objective of unconstrained reward maximization can lead to \textit{overoptimization} \citep{gao2022scaling}, that is, a policy that achieves high reward through exploitation of the idiosyncrasies of the reward function that are not aligned with the intended behavior. The most commonly-used objective in practice therefore combines reward maximization with a KL-divergence penalty between the language model and its initialization:
\begin{equation}
\label{eq:KLRL}
\max_{\pi_{\theta}}  \mathbb{E}_{x\sim \mathcal{D}_p, y\sim \pi_{\theta}(y \mid x)}\bigl[r(x, y) - \beta\log \frac{\pi_\theta(y \mid x)}{\pi_\text{ref}(y \mid x)}\bigr]
\end{equation}
where $\mathcal{D}_p$ is some dataset of prompts, $\piref$ is the reference model, usually the result of performing some supervised fine-tuning on a pre-trained model using demonstration data, and $\beta$ is a coefficient that controls the trade-off between reward and divergence \citep{ouyang2022training, bai2022training, stiennon2022learning}. Optimizing this objective aligns the model with the reward function without deviating too far from the pre-trained reference model, reducing overoptimization. In practice, the most common algorithm used to optimize this objective for language models is proximal policy optimization (PPO; \citet{schulman2017proximal}), although some variants exist \citep{Ramamurthy2022IsRL}. However, these algorithms are quite complex to implement and tune \citep{zheng2023secrets}.

\textbf{RL from preferences with direct preference optimization (DPO).} Most large language models fine-tuned with Eq.~\ref{eq:KLRL} optimize a reward function that is \textit{learned} from a dataset of preference rankings over possible model outputs. The DPO algorithm simplifies RL on language models for this special case \citep{rafailov2023direct}, using a dataset of preference pairs $\mathcal{D}=\{x^{(i)}, y_w^{(i)}, y_l^{(i)}\}_{i=1}^N$ of prompts $x$ and candidate responses $y_w$ and $y_l$ (typically sampled from $\pi_\text{ref}$), where $y_w$ is preferred over $y_l$ (denoted $y_w\succ y_l$). The probability of observing a particular preference pair is assumed to follow a Bradley-Terry model~\citep{bradley1952rank}:
\begin{equation}
    p(y_w\succ y_l)=\sigma(r(x, y_w)-r(x, y_l))
\end{equation}
where $\sigma$ is the sigmoid function and $r(x, y)$ is an unobserved reward or scoring function. \citet{rafailov2023direct} show that the optimal policy $\pi^*$ for the problem in Eq.~\ref{eq:KLRL} can be found by optimizing a simple classification loss computed directly on the preference data:
\begin{equation}
    \mathcal{L}_\text{DPO}(\pi_{\theta}; \piref) = -\mathbb{E}_{(x, y_w, y_l)\sim \mathcal{D}}\left[\log \sigma \left(\beta \log \frac{\pi_{\theta}(y_w\mid x)}{\piref(y_w\mid x)} - \beta \log \frac{\pi_{\theta}(y_l\mid x)}{\piref(y_l\mid x)}\right)\right]
\end{equation}
DPO enables learning $\pi_\theta$ from a fixed dataset of preferences, without fitting an explicit reward function or sampling from the policy in the loop of training (as is required in PPO). These advantages make DPO an attractive choice for fine-tuning language models for objectives other than imitation. However, a challenge remains in constructing preference pairs that encourage greater factuality.

\section{Constructing Preferences Encouraging Factuality in Long-Form Text}
\label{sec:pref-pair}
\begin{figure}
    \centering
    \includegraphics[width=\textwidth]{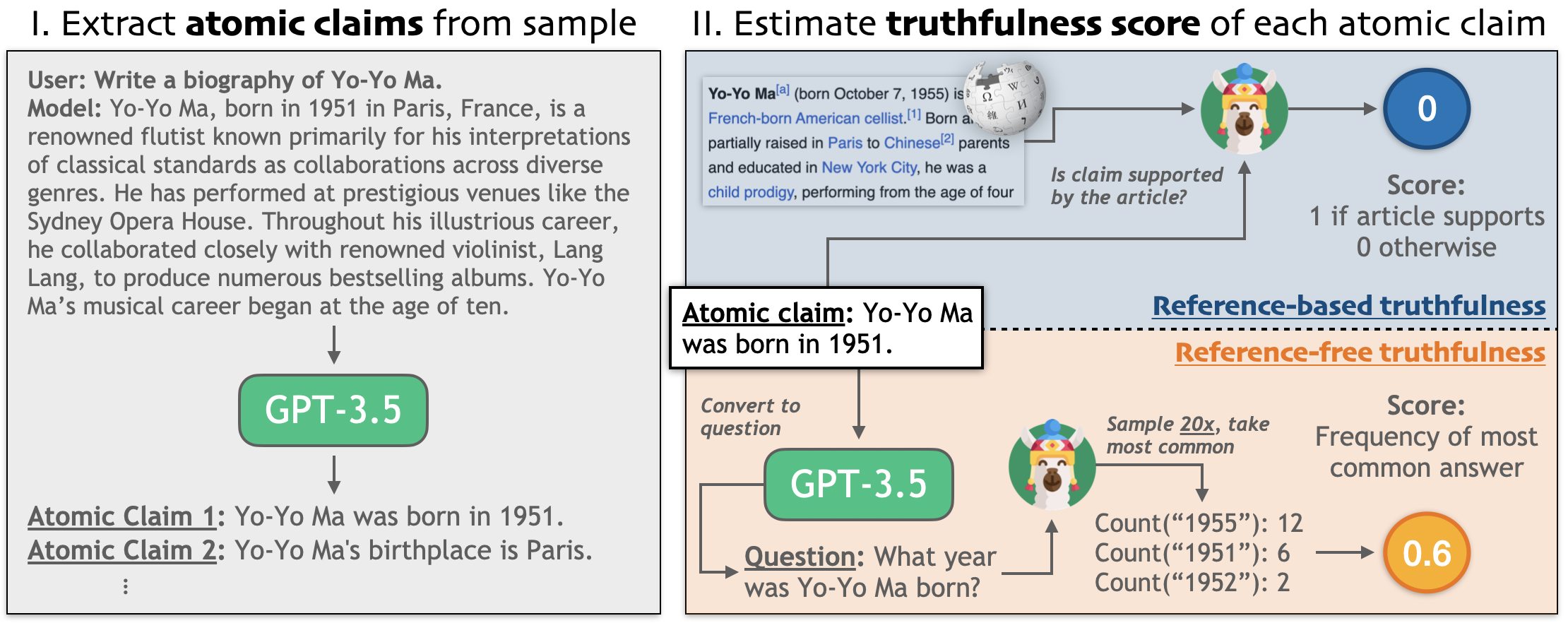}
    \caption{We estimate the factuality of a long-form generation by first extracting claims (left) and then evaluating the truthfulness of each claim (right). We consider two approaches for the latter: a \textit{reference-based} (top right) method that uses a fine-tuned Llama model to check if the fact is supported by Wikipedia \citep{min2023factscore}, and a \textit{reference-free} (bottom right) method that uses the model's confidence in its most likely answer to estimate its truthfulness.}
    \label{fig:method}
\end{figure}

While existing preference learning algorithms like DPO enable efficient, stable learning from objectives other than maximum likelihood, they require data in the form of preferences over possible responses to a prompt. In this section, we propose two classes of approaches to generating such preferences without human labeling effort. One class leverages existing methods to determine consistency with external reference texts as a measure of truthfulness; we propose another, which leverages calibrated model probabilities themselves as a proxy for truthfulness. For both approaches, we are computing an estimated \textbf{truthfulness score} over the claims in each generated response; the response with higher average truthfulness is taken as the preferred response. See Figure~\ref{fig:method} for an overview of both procedures for truthfulness scoring. Note that truthfulness scoring is needed \textbf{only at training time}; at test time, we can sample from the model in the normal manner.

\subsection{Reference-Based Truthfulness Estimation}
\label{sec:FS}
An intuitive approach to estimating truthfulness is by estimating the consistency of a given piece of text with a reliable reference text or knowledge base. Several recent works have introduced such evaluation criteria; for example, FactScore \citep{min2023factscore} uses Wikipedia as reference knowledge, and FacTool \citep{chern2023factool} uses Google Search Results. These measures show high agreement with human judgments of factuality, making them attractive sources of truth for preference data construction. Due to the relatively consistent and high quality of Wikipedia articles, we elect to use FactScore as a representative method of reference-based truthfulness scoring.

To evaluate a piece of text, FactScore first extracts a list of the atomic claims present in the text using GPT-3.5.\footnote{\url{https://platform.openai.com/docs/models/gpt-3-5}} For each atomic claim, a smaller, more efficient model such as a Llama-1-7B model \citep{touvron2023llama} that has been fine-tuned for fact-checking is then used to perform natural language inference \citep{MacCartneyManning08} to determine if a claim is supported by the reference text. The passage's truthfulness score is the fraction of the extracted atomic claims that are estimated to be supported by the reference text.

We note that reference-based truthfulness has the key limitation that it requires access to relevant, high-quality reference texts against which to measure consistency. Such a requirement may limit applicability to domains where ground truth documents are not known and accurate retrieval is difficult, such as in niche domains or less-structured tasks. Further, reference-based truthfulness estimation requires a reliable model to determine if an atomic claim is supported by the article. In light of these limitations, we propose a \textbf{reference-free} approach to estimating truthfulness of open-ended text, which avoids the need for retrieving external knowledge and checking consistency.

\subsection{Reference-Free Confidence-Based Truthfulness Estimation}
\label{sec:MC}
To eliminate the need for external knowledge, we leverage the fact that large language models are well-calibrated \citep{kadavath2022language,tian2023just}; that is, a large language model's confidence in a generated answer is highly correlated with the probability that the answer is correct. However, an open-ended passage might contain many facts, as well as particular stylistic choices that will have a significant impact on the total probability a model assigns to the text. Therefore, we first perform a claim extraction step, as in reference-based methods, and compute the average confidence of the model over all extracted factual claims as the final truthfulness score. The model used for computing confidence scores essentially takes the place of the reference text datastore.

More concretely, we first extract atomic claims from the text using GPT-3.5. We then use GPT-3.5 to convert each claim to a question testing knowledge of the particular fact. Careful rephrasing is necessary to ensure that the rephrased question is unambiguous; for example, the claim ``Yo-Yo Ma plays the cello'' should be converted to the question ``What instrument does Yo-Yo Ma play?'' rather than just ``What does Yo-Yo Ma play?'' as the latter question admits answers of the wrong type. If we were to use the second prompt, a model might assign 50\% of its probability on ``cello'' and 50\% of its probability on ``basketball.'' However, the model's low confidence is caused by the ambiguity of the question, \textit{not} low confidence in the instrument that Yo-Yo Ma plays. We detail the prompts used for question generation in Appendix \ref{sec:prompts}.

After each claim is converted to a minimally ambiguous question, we resample an answer 20 times, typically from the base model (e.g. Llama-1-7B) that is fine-tuned, to estimate the model's uncertainty over the answer. We use a few-shot prompt to encourage well-formed answers. We bin these answers by equivalence, using either heuristic string matching of the responses or using GPT-3.5 to assess if the answers are semantically equivalent, inspired by \citet{kuhn2023semantic}. Our heuristic string match checks whether the words in the answer, excluding stop words, are the same. We compare these choices in Section~\ref{sec:ablations}. The fraction of responses falling into the largest bin is the final truthfulness score used for the fact, essentially representing the model's confidence for this fact.

In Section~\ref{sec:ablations} we also evaluate a simpler approach to extracting atomic facts, by simply using named entities identified by a classifier \citep{spacy2}. This approach avoids using an external large language model for claim extraction and question rephrasing; instead, we simply resample the tokens in the original named entity in the response 20 times, binning them into buckets with equivalence checking, and again measure the fraction of responses falling into the largest bin as the confidence score.

\newif\ifincludeteaser
\includeteaserfalse

\ifdefined\arxiv
    \includeteasertrue
\fi
\ifdefined\neurips
    \includeteasertrue
\fi

\ifincludeteaser
    \definecolor{teaserblue}{rgb}{0.392, 0.584, 0.929}
    \begin{figure}
        \centering
        \includegraphics[width=0.85\textwidth]{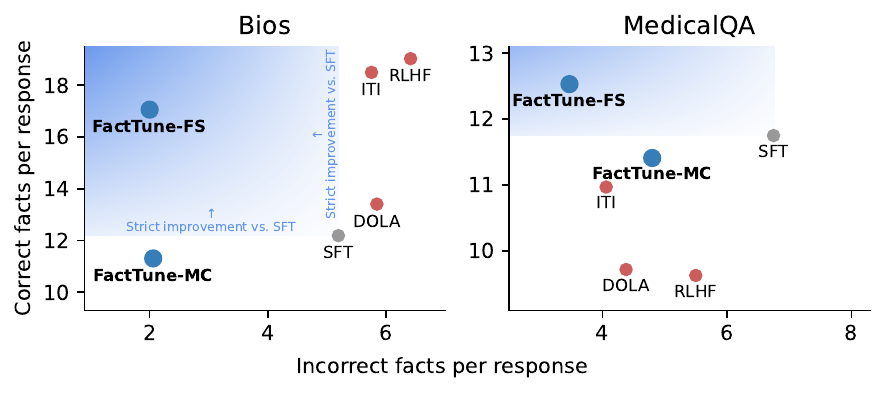}
        \caption{Factuality tuning (FactTune FS) is the only method that can produce a \textbf{strict improvement {\color{teaserblue}(shaded area)}} in factuality over the SFT model for the biography generation and medical question-answering problems. That is, only factuality tuning with FactScore-generated preferences (FS) simultaneously increases the number of correct statements and decreases the number of incorrect statements. Other approaches either increase the number of correct statements at the cost of more incorrect statements, or reduce the number of incorrect statements at the cost of fewer correct statements. Factuality tuning with model confidence-generated preferences (MC) lies just outside the strict improvement region.}
        \label{fig:teaser}
    \end{figure}
\fi

\subsection{Factuality Tuning: Putting it all Together}

Given a choice of truthfulness estimator, we can now construct a preference dataset for factuality tuning a given language model from a set of unlabeled prompts. First, we sample $n$ multiple candidate responses for each prompt from the model with simple temperature sampling with temperature 1.0 (using few-shot prompting for models that have not been fine-tuned). For each response, we then compute the truthfulness score with the chosen estimator (reference-based or reference-free). Finally, for all $\binom{n}{2}$ pairs of responses to each prompt, we simply choose the response with the higher truthfulness score as the preferred response. For a set of $m$ prompts, we ultimately generate $m\binom{n}{2}-k$ preference pairs, where $k$ is the number of pairs with equal scores. Finally, we fine-tune the model using the DPO pipeline, using all model responses as targets for the SFT stage.


\section{Experiments}

Our experiments evaluate the extent to which factuality can be learned through preference-based reinforcement learning, using the fully automated preference-generation pipeline described in Section \ref{sec:pref-pair}. We call the model fine-tuned with our reference-based metric FactTune-FS and the model fine-tuned with our model confidence-based score, which is completely reference-free, FactTune-MC. For all of our experiments, samples for model confidence are taken from Llama-1-7b.

\begin{table}
    \centering
    \small
    \resizebox{\textwidth}{!}{
    \begin{tabular}{ccccc}
        \toprule
        &  & \multirow{2}{*}{\shortstack{Prompts \\per Entity}} & \multirow{2}{*}{\shortstack{Responses \\per Prompt}} & Example prompt \\
        Dataset & Entities [train, test] & & & \\
        \midrule
        Biographies & 355 [296, 59] & 1 & 10 & Write a short biography of Mary Wollstonecraft. \\
        Medical QA & 200 [150, 50] & 6 & 6 & What are the common symptoms of a stroke? \\
        \bottomrule
    \end{tabular}
    }
    \caption{\textbf{Left.} Dataset statistics. In biographies, entities are individuals, and in MedicalQA, entities are medical conditions. We include 6 questions for each entity in MedicalQA, and we adjust the number of responses per prompt to keep the total number of pairs between thae two datasets roughly similar. \textbf{Right.} An example prompt from each dataset.}
    \label{tab:datasets}
\end{table}

\textbf{Datasets.} We conduct our experiments on two tasks: generating biographies and medical question-answering. For biographies, we generated a dataset consisting of 355 diverse well-known individuals (296 train, 59 test) with 10 short-paragraph biographies each. For medical question answering, we used a dataset of 200 diverse common medical conditions (150 train, 50 test) with 6 questions about each condition and 6 short-paragraph answers per question. The prompts were generated with GPT-3.5, and the answers were sampled from Llama-1-7B using a few-shot prompt for each dataset. We found that our procedure consistently resulted in well-formed and informative responses, albeit with possible factual errors. Because FactScore uses retrieval against a given Wikipedia article, we generate data based on individuals and medical conditions that have Wikipedia pages. 
See Table \ref{tab:datasets} for the summary stats and examples from our datasets.

\textbf{Baselines.} We compare factuality tuning with inference-time intervention \citep[ITI]{li2023inferencetime} and decoding by contrasting layers \citep[DOLA]{chuang2023dola}, applied to the SFT model for each task. For ITI, we supervise the training of the linear probes with FactScore labels: we take batches of atomic facts extracted from the training samples and bias the models' activations from the incorrect to correct atomic facts to determine the direction of the intervention. In the case of Llama-2, we also compare against `standard' RLHF with human preference labels \citep{touvron2023llama2}.

\textbf{Evaluation.} To evaluate each generated response, we follow the FactScore procedure to extract the number of correct and incorrect facts. Then, to check that the model responses are still relevant and helpful after actuality fine-tuning, we also use GPT-3.5 to determine whether each fact is relevant to the question or not (using the prompt in Appendix~\ref{sec:prompts}). For biographies, we observed that essentially 100\% of facts were relevant to the individual, so we skip the relevance computation to save costs. For each dataset, we report the number of correct and relevant facts (\# Correct), the number of inaccuracies (\# Incorrect), and the proportion of correct relevant facts out of the total number of extracted facts (\% Correct). Note that the total number of facts may vary between generations. We validate our evaluation metrics in Sec.~\ref{sec:validation}.

\subsection{Fine-Tuning for Factuality Across Domains}
\label{sec:main-exper}

\begin{table}
\small
\centering
\adjustbox{max width=\linewidth}{%
\begin{tabular}{@{}clcccccc@{}}
\toprule
 &  & \multicolumn{3}{c}{Biographies} & \multicolumn{3}{c}{Medical QA} \\ \cmidrule(lr){3-5} \cmidrule(lr){6-8}
Base Model & Method 
& \# Correct 
& \# Incorrect 
& \% Correct 
& \# Correct 
& \# Incorrect 
& \% Correct \\ \midrule
\multirow{5}{*}{Llama-1} & ITI & 11.67 &  \phantom{0}6.69 & 0.669 &  \phantom{0}8.91 & \phantom{0}5.16 & 0.633  \\
 & DOLA & 11.75 & \phantom{0}3.84 & 0.754 & \phantom{0}8.03 & \phantom{0}5.91 & 0.576 \\
 & SFT & 13.78 & 12.16 & 0.568 & 10.75 & \phantom{0}6.31 & 0.630 \\
 & FactTune-FS (ours) & \textbf{14.81} & \phantom{0}3.75 & \textbf{0.812} & 10.88 & \textbf{\phantom{0}4.50} & \textbf{0.707} \\
 & FactTune-MC (ours) & 10.59 & \textbf{\phantom{0}2.94} & 0.783 & \textbf{12.31} & \phantom{0}6.88 & 0.642 \\ \midrule
\multirow{6}{*}{Llama-2} & ITI & 18.50 & \phantom{0}5.75 & 0.760 & 10.97 & \phantom{0}4.06 & 0.730 \\
 & DOLA & 13.41 & \phantom{0}5.84 & 0.696 & \phantom{0}9.72 & \phantom{0}4.38 & 0.690 \\
 & Chat & \textbf{19.03} & \phantom{0}6.41 & 0.748 & \phantom{0}9.63 & \phantom{0}5.50 & 0.636 \\
 & SFT & 12.19 & \phantom{0}5.19 & 0.701 & 11.75 & \phantom{0}6.75 & 0.635 \\
 & FactTune-FS (ours) & 17.06 & \textbf{\phantom{0}2.00} & \textbf{0.895} & \textbf{12.53} & \textbf{\phantom{0}3.47} & \textbf{0.783} \\
 & FactTune-MC (ours) & 11.31 & \phantom{0}2.06 & 0.846 & 11.41 & \phantom{0}4.80 & 0.704 \\ \bottomrule
\end{tabular}
}
\caption{Factuality tuning from reference-based factuality-scored pairs (FactTune-FS) consistently improves factual accuracy compared to RLHF models and decoding-based factuality baselines, often reducing the number of factual errors \textit{and} increasing the number of correct facts generated. Factuality tuning from model-confidence scored pairs (FactTune-MC) also outperforms RLHF models and provides a strong reference-free alternate method for improving factuality and reducing error. }
\label{tab:main-table}
\end{table}



In this section, we apply our methodology for learning factuality to Llama-1-7b and Llama-2-7b in multiple domains. We show the results in Table \ref{tab:main-table}. Learning from reference-based factuality-scored pairs (FactTune-FS) consistently improves factual accuracy compared to RLHF models \textit{and} decoding-based factuality baselines by at least 23\% on biographies and 12\% on medical question-answering. FactTune-FS reduces the number of factual errors and maintains no more than a slight decrease, if not increase, in the amount of correct information generated. Factuality tuning from model-confidence scores (FactTune-MC) also reduces error rate and improves the factuality of RLHF models on both datasets, without any external reference information.



While our quantitative metrics demonstrate a clear increase in factual accuracy, we also wish to investigate how model generations change qualitatively after factuality fine-tuning. We observe that FactTune-FS and FactTune-MC samples tend to have more objective and direct sentences and less of a conversational or story-telling style compared to the SFT model (for example, see Appendix Table \ref{tab:examples-greta}). The  FactTune-FS and FactTune-MC samples have simpler sentences and lack casual phrases. As another example (in Appendix Table \ref{tab:examples}) the FactTune-FS and FactTune-MC biographies describe accurate facts, but not in a natural chronological order. GPT-4 rates FactTune-FS as less conversational in tone than the SFT model for 77.5\% (n=40) of Llama-1 questions and 65.6\% (n=32) of Llama-2 samples.

\subsection{Fine-tuning Chat Models for Factuality}
Most widely used practical chatbots today are LMs trained with RLHF to follow diverse instructions in a way that is helpful to users. In this section, we investigate the ability of our human-free factuality tuning method to improve the factuality of RLHF chat models. Using Llama-2-7b-Chat, we find that fine-tuning an RLHF LM with both factuality and semantic entropy-based rewards can further improve its factuality without significantly decreasing the total number of facts, as shown in Table \ref{tab:chat}.  
In other words, \textbf{factuality tuning can be composed with RLHF to further improve the factuality of chat models.} 

\begin{table}
\centering
\small
\adjustbox{max width=\linewidth}{%
\begin{tabular}{@{}clccccccc@{}}
\toprule
 &  & \multicolumn{3}{c}{Biographies} &  & \multicolumn{3}{c}{Medical QA} \\ \cmidrule(lr){3-5} \cmidrule(l){7-9} 
 Base Model  &  Method & \# Correct & \# Incorrect & \% Correct &  & \begin{tabular}[c]{@{}l@{}}\# Correct \end{tabular} & \# Incorrect & \begin{tabular}[c]{@{}l@{}}\% Correct \end{tabular} \\ \midrule
\multirow{4}{*}{Llama-2-Chat} & – & 19.03 & 6.41 & 0.748 &  & $\phantom{0}$9.63 & 5.50 & 0.636 \\
 & DOLA & \textbf{21.00} & 5.19 & 0.802 &  & \textbf{11.50} & 8.25 & 0.582 \\
 & FactTune-FS (ours) & 19.94 & \textbf{4.06} & \textbf{0.831} &  & $\phantom{0}$9.38 & \textbf{5.25} & \textbf{0.682} \\
 & FactTune-MC (ours) & 20.91 & 4.84 & 0.812 &  &  10.34 & 5.69 & 0.645 \\ \bottomrule
\end{tabular}
}
\caption{Factuality tuning a dialogue model (Llama-2-Chat) with both FactScore and model confidence-based truthfulness estimation (FactTune-FS, FactTune-MC) further improves its factual accuracy more than a baseline method for factuality, DOLA.}
\label{tab:chat}
\end{table}

\subsection{Complementary Benefits of Factuality Tuning and Decoding-Time Factuality Interventions}

Besides fine-tuning for factuality, multiple existing works aim to improve LLM factuality through inference time interventions to either the decoding process or the model parameters themselves. We explore the possibility of applying both of these types of methods together, i.e., using factuality-boosting decoding methods on a model fine-tuned with our factuality tuning procedure. In Table \ref{tab:DOLA_stack} we present the results of stacking both approaches. We find that in most cases, DOLA can even further increase the accuracy of factuality fine-tuned models, with one exception for Llama-2 on the biography task. While not a comprehensive evaluation of combining methods for improving factuality, this result suggests that different approaches to enhancing factuality may operate through complementary mechanisms.


\begin{table}
    \small
    \centering
    \adjustbox{max width=\linewidth}{%
    \begin{tabular}{clcccccc}
        \toprule
        & &  \multicolumn{3}{c}{Biographies} & \multicolumn{3}{c}{Medical QA} \\
        \cmidrule(lr){3-5} \cmidrule(lr){6-8} 
        Base Model & Method & \#Correct & \#Incorrect & \%Correct & \#Correct & \#Incorrect & \%Correct \\
        \midrule
        \multirow{2}{*}{Llama-1} & FactTune-FS & 14.81 & 3.75 & 0.812 & 10.88 & 4.50 & 0.707 \\
        & FactTune-FS + DOLA & 12.44 & 2.00 & 0.864 & 11.47 & 3.75 & 0.767 \\
        \midrule
        \multirow{2}{*}{Llama-2} &  FactTune-FS & 17.06 & 2.00 & 0.895 &  12.53 & 3.47 & 0.783 \\ 
        &  FactTune-FS + DOLA & 16.22 & 2.65 & 0.865 & 12.56 & 3.44 & 0.794 \\
        \bottomrule
    \end{tabular}
    }
    \caption{DOLA factuality decoding frequently composes with factuality fine-tuning, providing an increase in average correctness for the majority of combinations of model and dataset.} 
    \label{tab:DOLA_stack}
\end{table}

\subsection{Impact of Design Decisions of Open-Ended Model Confidence Scoring}
\label{sec:ablations}

We consider the impacts of different choices for each step in computing a reference-free truthfulness score for factuality tuning: fact extraction, confidence metric, and equivalence matching.


First, for the fact extraction step, we consider extracting questions about atomic facts identified by GPT-3.5 and sampling answers to each question, compared to extracting named entities for biographies, and noun chunks instead for Medical QA, using \texttt{nltk} and re-sampling the extracted entity. Atomic question extraction has the potential to be more comprehensive and precise, while named entity extraction is a less expensive proxy. In Table \ref{tab:SE_ablation}, we observe that atomic question extraction generally outperforms named entity extraction, although the difference in accuracy on the Medical QA dataset is small.

Next, we study the choice of confidence metric. The results in Table \ref{tab:SE_ablation} show that the choice of metric between maximum confidence---the probability of the largest semantic sample bin---or the entropy over the semantic bins varies, but maximum confidence provides a noticeable improvement to biographies under the atomic question setting. 

Finally, when binning samples, we consider replacing the heuristic equivalence match with an equivalence check by GPT-3.5. Surprisingly, using GPT-3.5 to determine equivalence between two samples produces \textit{worse-performing} preference pairs than using a simple string matching heuristic described in Section~\ref{sec:MC}. We suspect that this effect can potentially be caused by the following noise in GPT-3.5 equivalence checking: our heuristic equivalence match consistently underestimates semantic entropy across all examples, while GPT-3.5 matching could either over or underestimate samples, resulting in noisier preference pairs, even if GPT-3.5 equivalence check scores are closer to the true semantic entropy on average. GPT-4 could reduce this error, but we do not provide results due to its cost.

\begin{table}
    \small
    \centering
    \adjustbox{max width=\linewidth}{%
    \begin{tabular}{lllcccccc}
        \toprule
        & & & \multicolumn{3}{c}{Biographies} & \multicolumn{3}{c}{Medical QA} \\
        \cmidrule(lr){4-6} \cmidrule(lr){7-9} 
        Fact Ext. & Equiv & Metric & \#Correct & \#Incorrect & \%Correct & \#Correct & \#Incorrect & \%Correct \\
        \midrule
        \multirow{2}{*}{Entity} & \multirow{2}{*}{Heuristic}  & Entropy & 13.8 & 6.31 & 0.693  & 9.5 & 5.47 & 0.660 \\
                                & & Max Conf & 12.7 & 6.31 & 0.693 & 9.5 & 4.78 & 0.673 \\
        \midrule
        \multirow{2}{*}{Atomic} & \multirow{2}{*}{Heuristic}  & Entropy & 10.6 & 2.88 & 0.810 & 12.6 & 5.25 & 0.711 \\
                                & & Max Conf & 12.2 & 2.56 & 0.840 & 10.2 & 5.19 & 0.673 \\
        \midrule
        \multirow{2}{*}{Atomic} & \multirow{2}{*}{LLM}  & Entropy & 11.0 & 3.22 & 0.778 & 11.9 & 6.16 &  0.661 \\
                                & & Max Conf & 13.7 & 4.16 & 0.794 & 11.7 & 6.00 & 0.668 \\

        \bottomrule
    \end{tabular}
    }
    \caption{Model confidence-based preference construction with atomic question extraction during factuality scoring performs similarly or better than with named entity extraction. Surprisingly, using GPT-3.5 to determine equivalence between responses for semantic binning provides worse performance than a simple heuristic equivalence check. Note that we used 12 samples for all runs in this table.}
    \label{tab:SE_ablation}
\end{table}




\subsection{Validating Metrics for Factuality}
\label{sec:validation}
Our experiments primarily use counts of correct and incorrect facts computed by FactScore as the main evaluation metrics, as FactScore is automated and has been shown to exhibit good agreement with human fact-checkers \citep{min2023factscore}. Nonetheless, we aim to verify that our results are not specific or overfit to the FactScore criterion. In this section, we provide an evaluation with (1) human evaluators hired through Prolific.co\footnote{Human evaluators were compensated at an estimated hourly rate of \$16-18.} and (2) GPT-4. 

\begin{figure}
    \centering
    
    \begin{minipage}[b]{0.45\linewidth}
        \centering
        \small
        \adjustbox{max width=\linewidth}{%
        \begin{tabular}{llcc}
            \toprule
            Dataset & Evaluation & SFT & FactTune-FS \\
            \midrule
            Biographies & Human & 0.582 & 0.846 \\
            Biographies & FactScore & 0.669 & 0.921 \\
            MedQA & Human & 0.662 & 0.838 \\
            MedQA & FactScore & 0.534 & 0.806  \\
            \bottomrule
        \end{tabular}
        }
        \captionof{table}{To validate that our models do not suffer from extreme reward overoptimization, we conduct a human evaluation of the Llama-1-7B SFT and FactTune-FS models and find that an increase in FactScore also corresponds to a large increase in human-annotated accuracy.}
        \label{tab:human_eval}
    \end{minipage}
    \hfill
    \begin{minipage}[b]{0.5\linewidth}
        \centering
        \includegraphics[width=0.8\textwidth]{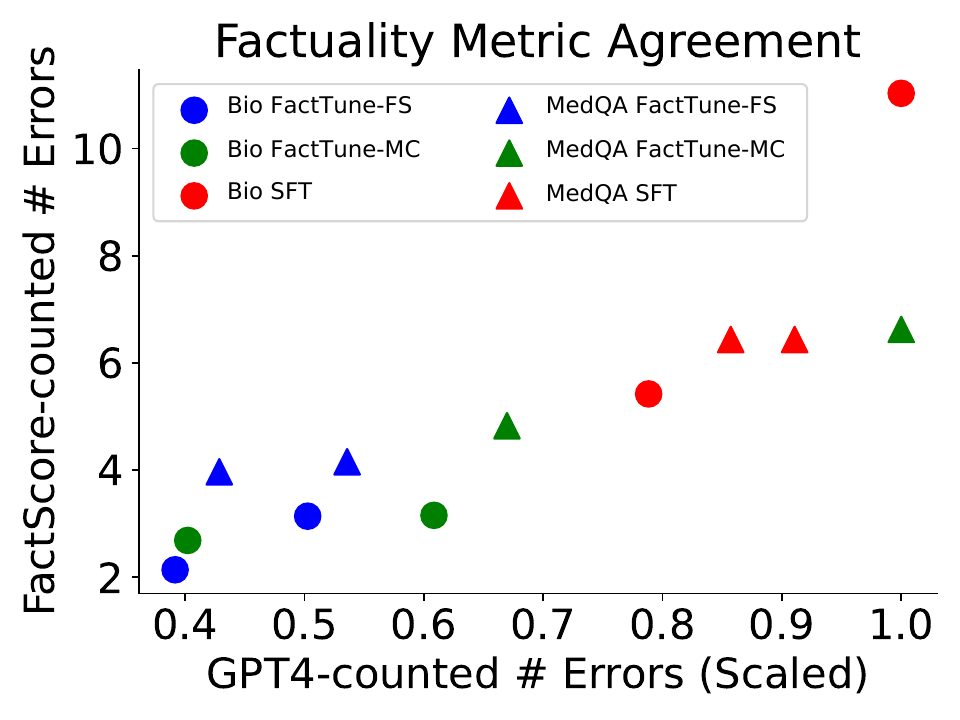}
        \caption{Average FactScore error counts and GPT-4 error counts are highly correlated, suggesting that the resulting models do not suffer from extreme reward overoptimization \citep{gao2022scaling}. We plot the average FactScore error count v.s. the average GPT-4-counted errors, scaling each dataset by the max GPT-4-error count in that dataset.}
        \label{fig:GPT-4}
    \end{minipage}
\end{figure}

To acquire human fact-checking results, we provide each human evaluator with a prompt, a generated response, and the title of the Wikipedia article they should use for fact-checking the response. We ask the human study participants to count the total number of facts and the number of incorrect facts in the response, and we divide these to obtain the human-rated accuracy. We provide the results in Table \ref{tab:human_eval}, where on average humans rated our FactTune-FS model for both datasets significantly higher than the SFT model.

Further, we ask GPT-4 to evaluate the factuality of a given response by counting the number of factual errors. We observe that the GPT-4 model ratings and FactScore model ratings are highly correlated, and GPT-4 provides another evaluation metric that demonstrates that FactTune-FS significantly reduces average error compared to the SFT models on both datasets (see Figure \ref{fig:GPT-4}). Taken together, these results suggest that the improvements in factuality are not the result of exploitation of our evaluation protocol.

\section{Related Work}

Many works have identified reducing factual errors (sometimes called `hallucinations') as a key challenge for building more reliable language models \citep{lewis2020retrieval,kadavath2022language,zhang2023language}, even for the most powerful language models \citep{bubeck2023sparks}. Other use of the term `hallucination' refers to summarization or translation system outputs not supported by the reference text \citep{maynez-etal-2020-faithfulness,zhang2020optimizing} even if they are factual \citep{cao-etal-2022-hallucinated}. Other work uses `hallucination' to describe vision-language models producing outputs not grounded in a visual input, e.g., a captioning system describing an object that doesn't exist in the image \citep{rohrbach-etal-2018-object}. In our case, we focus on statements that are factually incorrect (or, inconsistent with a set of `authoritative' texts, such as Wikipedia). 

Several works describe methods for detecting likely factual errors through sensitivity to perturbations in the prompt \citep{xu2023understanding}, high diversity of responses under resampling \citep{kadavath2022language,mündler2023selfcontradictory,kuhn2023semantic}, or inconsistency with external knowledge sources \citep{min2023factscore,chern2023factool}, or properties of internal activations \citep{azaria2023internal}. Others go beyond detecting errors, correcting them after they have been generated \citep{peng2023check,gao2023rarr,dhuliawala2023chainofverification}. These approaches typically rely on retrieving relevant data from a trusted knowledge base and use another LLM to verify consistency; however, retrieval-based methods face key challenges, namely reliable resolution of conflicts between parametric and retrieved knowledge \citep{longpre2022entitybased,chen-etal-2022-rich} as well as maintaining improvements in factuality as model size increases \citep{mallen-etal-2023-trust}. Further, retrieval-based methods add significant system complexity; the most common open-source consumer language models thus use purely parametric models \citep{touvron2023llama}. The FactScore variant of our approach uses retrieval only during training, avoiding inference time complexity.

Most similar to ours, some approaches attempt to prevent the generation of factual errors in the first place, using prompting strategies \citep{si2023prompting} or perturbing the internal representations of the model \citep{chuang2023dola, li2023inferencetime}. Unlike using a fixed heuristic for identifying an internal `factuality' dimension, we optimize directly for the end goal of generating factual statements, which we find shows a greater improvement in factuality. Finally, while most past work has focused on short-form NLG tasks like short-form question-answering \citep{kadavath2022language}, we explore ways to measure model confidence over factual information in long-form, unstructured text and estimate truthfulness in a reference-free manner (i.e., don't require any external knowledge base or annotations).

\section{Conclusion}
In this paper, we show a practical, effective strategy to improve a language model's ability to generate factual content, specifically focusing on long-form generations. We develop and study two different approaches to estimating the truthfulness of long-form text and optimize for these criteria using preference-based learning. In addition to existing \textit{reference-based} truthfulness estimators that leverage external knowledge to establish the truth of a particular statement, we introduce a novel \textit{reference-free} procedure for estimating truthfulness that uses the language model's own uncertainty as an indication of factuality. Our experiments show that fine-tuning a language model with either criterion reliably reduces the number of incorrect facts (i.e. hallucinations) that the model generates. Reference-free approaches like the one we have introduced provide a particularly scalable self-supervision strategy to improve factuality, eliminating the need for a reference corpus of `gold' texts.

The experimental results suggest a number of avenues for future work. First, because of the limited research and thus the limited benchmarks on the factuality of long-form language model generations, we proposed two new tasks to benchmark our approach. These tasks are representative of but do not fully cover the range of scenarios where we would hope to improve factuality. Furthermore, our experiments provide evidence for improving the factuality of dialogue models that are already fine-tuned with RLHF, but still leave open the question of how best to combine typical RLHF rewards and approaches with factuality rankings. Similarly, exploring additional ways to combine factuality tuning with existing methods for improving factuality, such as in our factuality tuning + DOLA experiment, may be a fruitful direction for future research. Finally, we explore only 7B models in this work. Scaling up our factuality tuning recipe to larger models (and larger preference datasets) may reduce hallucinations even further.

\ifdefined\iclr
\textbf{Reproducibility Statement.} We explain the steps of our fine-tuning method in Section \ref{sec:pref-pair}. In Section \ref{sec:main-exper}, we provide details on the dataset (dataset statistics, how it was generated, and examples), as well as how the evaluation is completed and how we implemented the baselines. In the experiment subsections and captions, we provide additional implementation or reporting details. In the appendix, we provide the exact GPT-3.5 prompts used for the extraction steps of our reference-free scoring method. An anonymized implementation of our experiments can be provided during the review period.
\fi
\section*{Acknowledgements}
EM gratefully acknowledges funding from a Knight-Hennessy graduate fellowship and a Stanford Accelerator for Generative AI and Education grant. CF and CDM are CIFAR Fellows.

\clearpage
\bibliography{main}

\begin{thebibliography}{42}
\providecommand{\natexlab}[1]{#1}
\providecommand{\url}[1]{\texttt{#1}}
\expandafter\ifx\csname urlstyle\endcsname\relax
  \providecommand{\doi}[1]{doi: #1}\else
  \providecommand{\doi}{doi: \begingroup \urlstyle{rm}\Url}\fi

\bibitem[Agrawal et~al.(2023)Agrawal, Suzgun, Mackey, and Kalai]{agrawal2023language}
Ayush Agrawal, Mirac Suzgun, Lester Mackey, and Adam~Tauman Kalai.
\newblock Do language models know when they're hallucinating references?, 2023.
\newblock arXiv preprint arxiv:2305.18248.

\bibitem[Azaria \& Mitchell(2023)Azaria and Mitchell]{azaria2023internal}
Amos Azaria and Tom Mitchell.
\newblock The internal state of an {LLM} knows when its lying, 2023.

\bibitem[Bai et~al.(2022)Bai, Jones, Ndousse, Askell, Chen, DasSarma, Drain, Fort, Ganguli, Henighan, Joseph, Kadavath, Kernion, Conerly, El-Showk, Elhage, Hatfield-Dodds, Hernandez, Hume, Johnston, Kravec, Lovitt, Nanda, Olsson, Amodei, Brown, Clark, McCandlish, Olah, Mann, and Kaplan]{bai2022training}
Yuntao Bai, Andy Jones, Kamal Ndousse, Amanda Askell, Anna Chen, Nova DasSarma, Dawn Drain, Stanislav Fort, Deep Ganguli, Tom Henighan, Nicholas Joseph, Saurav Kadavath, Jackson Kernion, Tom Conerly, Sheer El-Showk, Nelson Elhage, Zac Hatfield-Dodds, Danny Hernandez, Tristan Hume, Scott Johnston, Shauna Kravec, Liane Lovitt, Neel Nanda, Catherine Olsson, Dario Amodei, Tom Brown, Jack Clark, Sam McCandlish, Chris Olah, Ben Mann, and Jared Kaplan.
\newblock Training a helpful and harmless assistant with reinforcement learning from human feedback, 2022.

\bibitem[Bradley \& Terry(1952)Bradley and Terry]{bradley1952rank}
Ralph~Allan Bradley and Milton~E Terry.
\newblock Rank analysis of incomplete block designs: I. the method of paired comparisons.
\newblock \emph{Biometrika}, 39\penalty0 (3/4):\penalty0 324--345, 1952.

\bibitem[Bubeck et~al.(2023)Bubeck, Chandrasekaran, Eldan, Gehrke, Horvitz, Kamar, Lee, Lee, Li, Lundberg, Nori, Palangi, Ribeiro, and Zhang]{bubeck2023sparks}
Sébastien Bubeck, Varun Chandrasekaran, Ronen Eldan, Johannes Gehrke, Eric Horvitz, Ece Kamar, Peter Lee, Yin~Tat Lee, Yuanzhi Li, Scott Lundberg, Harsha Nori, Hamid Palangi, Marco~Tulio Ribeiro, and Yi~Zhang.
\newblock Sparks of artificial general intelligence: Early experiments with gpt-4, 2023.

\bibitem[Cao et~al.(2022)Cao, Dong, and Cheung]{cao-etal-2022-hallucinated}
Meng Cao, Yue Dong, and Jackie Cheung.
\newblock Hallucinated but factual! inspecting the factuality of hallucinations in abstractive summarization.
\newblock In \emph{Proceedings of the 60th Annual Meeting of the Association for Computational Linguistics (Volume 1: Long Papers)}, pp.\  3340--3354, Dublin, Ireland, May 2022. Association for Computational Linguistics.
\newblock \doi{10.18653/v1/2022.acl-long.236}.
\newblock URL \url{https://aclanthology.org/2022.acl-long.236}.

\bibitem[Chen et~al.(2022)Chen, Zhang, and Choi]{chen-etal-2022-rich}
Hung-Ting Chen, Michael Zhang, and Eunsol Choi.
\newblock Rich knowledge sources bring complex knowledge conflicts: Recalibrating models to reflect conflicting evidence.
\newblock In \emph{Proceedings of the 2022 Conference on Empirical Methods in Natural Language Processing}, pp.\  2292--2307, Abu Dhabi, United Arab Emirates, December 2022. Association for Computational Linguistics.
\newblock \doi{10.18653/v1/2022.emnlp-main.146}.
\newblock URL \url{https://aclanthology.org/2022.emnlp-main.146}.

\bibitem[Chen et~al.(2021)Chen, Tworek, Jun, Yuan, Ponde, Kaplan, Edwards, Burda, Joseph, Brockman, Ray, Puri, Krueger, Petrov, Khlaaf, Sastry, Mishkin, Chan, Gray, Ryder, Pavlov, Power, Kaiser, Bavarian, Winter, Tillet, Such, Cummings, Plappert, Chantzis, Barnes, Herbert-Voss, Guss, Nichol, Babuschkin, Balaji, Jain, Carr, Leike, Achiam, Misra, Morikawa, Radford, Knight, Brundage, Murati, Mayer, Welinder, McGrew, Amodei, McCandlish, Sutskever, and Zaremba]{Chen2021EvaluatingLL}
Mark Chen, Jerry Tworek, Heewoo Jun, Qiming Yuan, Henrique Ponde, Jared Kaplan, Harrison Edwards, Yura Burda, Nicholas Joseph, Greg Brockman, Alex Ray, Raul Puri, Gretchen Krueger, Michael Petrov, Heidy Khlaaf, Girish Sastry, Pamela Mishkin, Brooke Chan, Scott Gray, Nick Ryder, Mikhail Pavlov, Alethea Power, Lukasz Kaiser, Mohammad Bavarian, Clemens Winter, Philippe Tillet, Felipe~Petroski Such, David~W. Cummings, Matthias Plappert, Fotios Chantzis, Elizabeth Barnes, Ariel Herbert-Voss, William~H. Guss, Alex Nichol, Igor Babuschkin, S.~Arun Balaji, Shantanu Jain, Andrew Carr, Jan Leike, Joshua Achiam, Vedant Misra, Evan Morikawa, Alec Radford, Matthew~M. Knight, Miles Brundage, Mira Murati, Katie Mayer, Peter Welinder, Bob McGrew, Dario Amodei, Sam McCandlish, Ilya Sutskever, and Wojciech Zaremba.
\newblock Evaluating large language models trained on code.
\newblock \emph{ArXiv}, abs/2107.03374, 2021.
\newblock URL \url{https://api.semanticscholar.org/CorpusID:235755472}.

\bibitem[Chern et~al.(2023)Chern, Chern, Chen, Yuan, Feng, Zhou, He, Neubig, and Liu]{chern2023factool}
I-Chun Chern, Steffi Chern, Shiqi Chen, Weizhe Yuan, Kehua Feng, Chunting Zhou, Junxian He, Graham Neubig, and Pengfei Liu.
\newblock Factool: Factuality detection in generative ai -- a tool augmented framework for multi-task and multi-domain scenarios, 2023.

\bibitem[Christiano et~al.(2017)Christiano, Leike, Brown, Martic, Legg, and Amodei]{christiano2017deep}
Paul~F Christiano, Jan Leike, Tom Brown, Miljan Martic, Shane Legg, and Dario Amodei.
\newblock Deep reinforcement learning from human preferences.
\newblock In I.~Guyon, U.~Von Luxburg, S.~Bengio, H.~Wallach, R.~Fergus, S.~Vishwanathan, and R.~Garnett (eds.), \emph{Advances in Neural Information Processing Systems}, volume~30. Curran Associates, Inc., 2017.
\newblock URL \url{https://proceedings.neurips.cc/paper_files/paper/2017/file/d5e2c0adad503c91f91df240d0cd4e49-Paper.pdf}.

\bibitem[Chuang et~al.(2023)Chuang, Xie, Luo, Kim, Glass, and He]{chuang2023dola}
Yung-Sung Chuang, Yujia Xie, Hongyin Luo, Yoon Kim, James Glass, and Pengcheng He.
\newblock Dola: Decoding by contrasting layers improves factuality in large language models, 2023.

\bibitem[Chung et~al.(2022)Chung, Hou, Longpre, Zoph, Tay, Fedus, Li, Wang, Dehghani, Brahma, Webson, Gu, Dai, Suzgun, Chen, Chowdhery, Castro-Ros, Pellat, Robinson, Valter, Narang, Mishra, Yu, Zhao, Huang, Dai, Yu, Petrov, Chi, Dean, Devlin, Roberts, Zhou, Le, and Wei]{chung2022scaling}
Hyung~Won Chung, Le~Hou, Shayne Longpre, Barret Zoph, Yi~Tay, William Fedus, Yunxuan Li, Xuezhi Wang, Mostafa Dehghani, Siddhartha Brahma, Albert Webson, Shixiang~Shane Gu, Zhuyun Dai, Mirac Suzgun, Xinyun Chen, Aakanksha Chowdhery, Alex Castro-Ros, Marie Pellat, Kevin Robinson, Dasha Valter, Sharan Narang, Gaurav Mishra, Adams Yu, Vincent Zhao, Yanping Huang, Andrew Dai, Hongkun Yu, Slav Petrov, Ed~H. Chi, Jeff Dean, Jacob Devlin, Adam Roberts, Denny Zhou, Quoc~V. Le, and Jason Wei.
\newblock Scaling instruction-finetuned language models, 2022.

\bibitem[Dhuliawala et~al.(2023)Dhuliawala, Komeili, Xu, Raileanu, Li, Celikyilmaz, and Weston]{dhuliawala2023chainofverification}
Shehzaad Dhuliawala, Mojtaba Komeili, Jing Xu, Roberta Raileanu, Xian Li, Asli Celikyilmaz, and Jason Weston.
\newblock Chain-of-verification reduces hallucination in large language models, 2023.

\bibitem[Gao et~al.(2022)Gao, Schulman, and Hilton]{gao2022scaling}
Leo Gao, John Schulman, and Jacob Hilton.
\newblock Scaling laws for reward model overoptimization, 2022.

\bibitem[Gao et~al.(2023)Gao, Dai, Pasupat, Chen, Chaganty, Fan, Zhao, Lao, Lee, Juan, and Guu]{gao2023rarr}
Luyu Gao, Zhuyun Dai, Panupong Pasupat, Anthony Chen, Arun~Tejasvi Chaganty, Yicheng Fan, Vincent~Y. Zhao, Ni~Lao, Hongrae Lee, Da-Cheng Juan, and Kelvin Guu.
\newblock Rarr: Researching and revising what language models say, using language models, 2023.

\bibitem[Honnibal \& Montani(2017)Honnibal and Montani]{spacy2}
Matthew Honnibal and Ines Montani.
\newblock {spaCy 2}: Natural language understanding with {B}loom embeddings, convolutional neural networks and incremental parsing.
\newblock To appear, 2017.

\bibitem[Kadavath et~al.(2022)Kadavath, Conerly, Askell, Henighan, Drain, Perez, Schiefer, Hatfield-Dodds, DasSarma, Tran-Johnson, Johnston, El-Showk, Jones, Elhage, Hume, Chen, Bai, Bowman, Fort, Ganguli, Hernandez, Jacobson, Kernion, Kravec, Lovitt, Ndousse, Olsson, Ringer, Amodei, Brown, Clark, Joseph, Mann, McCandlish, Olah, and Kaplan]{kadavath2022language}
Saurav Kadavath, Tom Conerly, Amanda Askell, Tom Henighan, Dawn Drain, Ethan Perez, Nicholas Schiefer, Zac Hatfield-Dodds, Nova DasSarma, Eli Tran-Johnson, Scott Johnston, Sheer El-Showk, Andy Jones, Nelson Elhage, Tristan Hume, Anna Chen, Yuntao Bai, Sam Bowman, Stanislav Fort, Deep Ganguli, Danny Hernandez, Josh Jacobson, Jackson Kernion, Shauna Kravec, Liane Lovitt, Kamal Ndousse, Catherine Olsson, Sam Ringer, Dario Amodei, Tom Brown, Jack Clark, Nicholas Joseph, Ben Mann, Sam McCandlish, Chris Olah, and Jared Kaplan.
\newblock Language models (mostly) know what they know, 2022.
\newblock URL \url{http://arxiv.org/abs/2207.05221}.
\newblock Arxiv arxiv:2207.05221.

\bibitem[Kuhn et~al.(2023)Kuhn, Gal, and Farquhar]{kuhn2023semantic}
Lorenz Kuhn, Yarin Gal, and Sebastian Farquhar.
\newblock Semantic uncertainty: Linguistic invariances for uncertainty estimation in natural language generation, 2023.

\bibitem[Lewis et~al.(2020)Lewis, Perez, Piktus, Petroni, Karpukhin, Goyal, K\"{u}ttler, Lewis, Yih, Rockt\"{a}schel, Riedel, and Kiela]{lewis2020retrieval}
Patrick Lewis, Ethan Perez, Aleksandra Piktus, Fabio Petroni, Vladimir Karpukhin, Naman Goyal, Heinrich K\"{u}ttler, Mike Lewis, Wen-tau Yih, Tim Rockt\"{a}schel, Sebastian Riedel, and Douwe Kiela.
\newblock Retrieval-augmented generation for knowledge-intensive {NLP} tasks.
\newblock In H.~Larochelle, M.~Ranzato, R.~Hadsell, M.F. Balcan, and H.~Lin (eds.), \emph{Advances in Neural Information Processing Systems}, volume~33, pp.\  9459--9474. Curran Associates, Inc., 2020.
\newblock URL \url{https://proceedings.neurips.cc/paper_files/paper/2020/file/6b493230205f780e1bc26945df7481e5-Paper.pdf}.

\bibitem[Li et~al.(2023)Li, Patel, Viégas, Pfister, and Wattenberg]{li2023inferencetime}
Kenneth Li, Oam Patel, Fernanda Viégas, Hanspeter Pfister, and Martin Wattenberg.
\newblock Inference-time intervention: Eliciting truthful answers from a language model, 2023.

\bibitem[Longpre et~al.(2022)Longpre, Perisetla, Chen, Ramesh, DuBois, and Singh]{longpre2022entitybased}
Shayne Longpre, Kartik Perisetla, Anthony Chen, Nikhil Ramesh, Chris DuBois, and Sameer Singh.
\newblock Entity-based knowledge conflicts in question answering, 2022.

\bibitem[MacCartney \& Manning(2008)MacCartney and Manning]{MacCartneyManning08}
Bill MacCartney and Christopher~D. Manning.
\newblock Modeling semantic containment and exclusion in natural language inference.
\newblock In \emph{Proceedings of the 22nd International Conference on Computational Linguistics (Coling 2008)}, pp.\  521--528, Manchester, UK, August 2008. Coling 2008 Organizing Committee.
\newblock URL \url{http://www.aclweb.org/anthology/C08-1066}.

\bibitem[Mallen et~al.(2023)Mallen, Asai, Zhong, Das, Khashabi, and Hajishirzi]{mallen-etal-2023-trust}
Alex Mallen, Akari Asai, Victor Zhong, Rajarshi Das, Daniel Khashabi, and Hannaneh Hajishirzi.
\newblock When not to trust language models: Investigating effectiveness of parametric and non-parametric memories.
\newblock In \emph{Proceedings of the 61st Annual Meeting of the Association for Computational Linguistics (Volume 1: Long Papers)}, pp.\  9802--9822, Toronto, Canada, July 2023. Association for Computational Linguistics.
\newblock \doi{10.18653/v1/2023.acl-long.546}.
\newblock URL \url{https://aclanthology.org/2023.acl-long.546}.

\bibitem[Maynez et~al.(2020)Maynez, Narayan, Bohnet, and McDonald]{maynez-etal-2020-faithfulness}
Joshua Maynez, Shashi Narayan, Bernd Bohnet, and Ryan McDonald.
\newblock On faithfulness and factuality in abstractive summarization.
\newblock In \emph{Proceedings of the 58th Annual Meeting of the Association for Computational Linguistics}, pp.\  1906--1919, Online, July 2020. Association for Computational Linguistics.
\newblock \doi{10.18653/v1/2020.acl-main.173}.
\newblock URL \url{https://aclanthology.org/2020.acl-main.173}.

\bibitem[Min et~al.(2023)Min, Krishna, Lyu, Lewis, tau Yih, Koh, Iyyer, Zettlemoyer, and Hajishirzi]{min2023factscore}
Sewon Min, Kalpesh Krishna, Xinxi Lyu, Mike Lewis, Wen tau Yih, Pang~Wei Koh, Mohit Iyyer, Luke Zettlemoyer, and Hannaneh Hajishirzi.
\newblock Factscore: Fine-grained atomic evaluation of factual precision in long form text generation, 2023.

\bibitem[Mündler et~al.(2023)Mündler, He, Jenko, and Vechev]{mündler2023selfcontradictory}
Niels Mündler, Jingxuan He, Slobodan Jenko, and Martin Vechev.
\newblock Self-contradictory hallucinations of large language models: Evaluation, detection and mitigation, 2023.

\bibitem[Ouyang et~al.(2022)Ouyang, Wu, Jiang, Almeida, Wainwright, Mishkin, Zhang, Agarwal, Slama, Ray, Schulman, Hilton, Kelton, Miller, Simens, Askell, Welinder, Christiano, Leike, and Lowe]{ouyang2022training}
Long Ouyang, Jeff Wu, Xu~Jiang, Diogo Almeida, Carroll~L. Wainwright, Pamela Mishkin, Chong Zhang, Sandhini Agarwal, Katarina Slama, Alex Ray, John Schulman, Jacob Hilton, Fraser Kelton, Luke Miller, Maddie Simens, Amanda Askell, Peter Welinder, Paul Christiano, Jan Leike, and Ryan Lowe.
\newblock Training language models to follow instructions with human feedback, 2022.

\bibitem[Peng et~al.(2023)Peng, Galley, He, Cheng, Xie, Hu, Huang, Liden, Yu, Chen, and Gao]{peng2023check}
Baolin Peng, Michel Galley, Pengcheng He, Hao Cheng, Yujia Xie, Yu~Hu, Qiuyuan Huang, Lars Liden, Zhou Yu, Weizhu Chen, and Jianfeng Gao.
\newblock Check your facts and try again: Improving large language models with external knowledge and automated feedback, 2023.

\bibitem[Rafailov et~al.(2023)Rafailov, Sharma, Mitchell, Ermon, Manning, and Finn]{rafailov2023direct}
Rafael Rafailov, Archit Sharma, Eric Mitchell, Stefano Ermon, Christopher~D. Manning, and Chelsea Finn.
\newblock Direct preference optimization: Your language model is secretly a reward model, 2023.

\bibitem[Ramamurthy et~al.(2022)Ramamurthy, Ammanabrolu, Brantley, Hessel, Sifa, Bauckhage, Hajishirzi, and Choi]{Ramamurthy2022IsRL}
Rajkumar Ramamurthy, Prithviraj Ammanabrolu, Kiant{\'e} Brantley, Jack Hessel, Rafet Sifa, Christian Bauckhage, Hannaneh Hajishirzi, and Yejin Choi.
\newblock Is reinforcement learning (not) for natural language processing?: Benchmarks, baselines, and building blocks for natural language policy optimization.
\newblock 2022.
\newblock URL \url{https://arxiv.org/abs/2210.01241}.

\bibitem[Rohrbach et~al.(2018)Rohrbach, Hendricks, Burns, Darrell, and Saenko]{rohrbach-etal-2018-object}
Anna Rohrbach, Lisa~Anne Hendricks, Kaylee Burns, Trevor Darrell, and Kate Saenko.
\newblock Object hallucination in image captioning.
\newblock In Ellen Riloff, David Chiang, Julia Hockenmaier, and Jun{'}ichi Tsujii (eds.), \emph{Proceedings of the 2018 Conference on Empirical Methods in Natural Language Processing}, pp.\  4035--4045, Brussels, Belgium, October-November 2018. Association for Computational Linguistics.
\newblock \doi{10.18653/v1/D18-1437}.
\newblock URL \url{https://aclanthology.org/D18-1437}.

\bibitem[Schulman et~al.(2017)Schulman, Wolski, Dhariwal, Radford, and Klimov]{schulman2017proximal}
John Schulman, Filip Wolski, Prafulla Dhariwal, Alec Radford, and Oleg Klimov.
\newblock Proximal policy optimization algorithms, 2017.

\bibitem[Si et~al.(2023)Si, Gan, Yang, Wang, Wang, Boyd-Graber, and Wang]{si2023prompting}
Chenglei Si, Zhe Gan, Zhengyuan Yang, Shuohang Wang, Jianfeng Wang, Jordan Boyd-Graber, and Lijuan Wang.
\newblock Prompting gpt-3 to be reliable, 2023.

\bibitem[Stiennon et~al.(2020)Stiennon, Ouyang, Wu, Ziegler, Lowe, Voss, Radford, Amodei, and Christiano]{stiennon2022learning}
Nisan Stiennon, Long Ouyang, Jeff Wu, Daniel~M. Ziegler, Ryan Lowe, Chelsea Voss, Alec Radford, Dario Amodei, and Paul Christiano.
\newblock Learning to summarize from human feedback.
\newblock \emph{Neural Information Processing Systems}, 18, 2020.

\bibitem[Tian et~al.(2023)Tian, Mitchell, Zhou, Sharma, Rafailov, Yao, Finn, and Manning]{tian2023just}
Katherine Tian, Eric Mitchell, Allan Zhou, Archit Sharma, Rafael Rafailov, Huaxiu Yao, Chelsea Finn, and Christopher~D. Manning.
\newblock Just ask for calibration: Strategies for eliciting calibrated confidence scores from language models fine-tuned with human feedback, 2023.

\bibitem[Touvron et~al.(2023{\natexlab{a}})Touvron, Lavril, Izacard, Martinet, Lachaux, Lacroix, Rozière, Goyal, Hambro, Azhar, Rodriguez, Joulin, Grave, and Lample]{touvron2023llama}
Hugo Touvron, Thibaut Lavril, Gautier Izacard, Xavier Martinet, Marie-Anne Lachaux, Timothée Lacroix, Baptiste Rozière, Naman Goyal, Eric Hambro, Faisal Azhar, Aurelien Rodriguez, Armand Joulin, Edouard Grave, and Guillaume Lample.
\newblock Llama: Open and efficient foundation language models, 2023{\natexlab{a}}.

\bibitem[Touvron et~al.(2023{\natexlab{b}})Touvron, Martin, Stone, Albert, Almahairi, Babaei, Bashlykov, Batra, Bhargava, Bhosale, Bikel, Blecher, Ferrer, Chen, Cucurull, Esiobu, Fernandes, Fu, Fu, Fuller, Gao, Goswami, Goyal, Hartshorn, Hosseini, Hou, Inan, Kardas, Kerkez, Khabsa, Kloumann, Korenev, Koura, Lachaux, Lavril, Lee, Liskovich, Lu, Mao, Martinet, Mihaylov, Mishra, Molybog, Nie, Poulton, Reizenstein, Rungta, Saladi, Schelten, Silva, Smith, Subramanian, Tan, Tang, Taylor, Williams, Kuan, Xu, Yan, Zarov, Zhang, Fan, Kambadur, Narang, Rodriguez, Stojnic, Edunov, and Scialom]{touvron2023llama2}
Hugo Touvron, Louis Martin, Kevin Stone, Peter Albert, Amjad Almahairi, Yasmine Babaei, Nikolay Bashlykov, Soumya Batra, Prajjwal Bhargava, Shruti Bhosale, Dan Bikel, Lukas Blecher, Cristian~Canton Ferrer, Moya Chen, Guillem Cucurull, David Esiobu, Jude Fernandes, Jeremy Fu, Wenyin Fu, Brian Fuller, Cynthia Gao, Vedanuj Goswami, Naman Goyal, Anthony Hartshorn, Saghar Hosseini, Rui Hou, Hakan Inan, Marcin Kardas, Viktor Kerkez, Madian Khabsa, Isabel Kloumann, Artem Korenev, Punit~Singh Koura, Marie-Anne Lachaux, Thibaut Lavril, Jenya Lee, Diana Liskovich, Yinghai Lu, Yuning Mao, Xavier Martinet, Todor Mihaylov, Pushkar Mishra, Igor Molybog, Yixin Nie, Andrew Poulton, Jeremy Reizenstein, Rashi Rungta, Kalyan Saladi, Alan Schelten, Ruan Silva, Eric~Michael Smith, Ranjan Subramanian, Xiaoqing~Ellen Tan, Binh Tang, Ross Taylor, Adina Williams, Jian~Xiang Kuan, Puxin Xu, Zheng Yan, Iliyan Zarov, Yuchen Zhang, Angela Fan, Melanie Kambadur, Sharan Narang, Aurelien Rodriguez, Robert Stojnic, Sergey Edunov, and Thomas
  Scialom.
\newblock Llama 2: Open foundation and fine-tuned chat models, 2023{\natexlab{b}}.

\bibitem[Xu et~al.(2023)Xu, Agrawal, Briakou, Martindale, and Carpuat]{xu2023understanding}
Weijia Xu, Sweta Agrawal, Eleftheria Briakou, Marianna~J. Martindale, and Marine Carpuat.
\newblock {Understanding and Detecting Hallucinations in Neural Machine Translation via Model Introspection}.
\newblock \emph{Transactions of the Association for Computational Linguistics}, 11:\penalty0 546--564, 06 2023.
\newblock ISSN 2307-387X.
\newblock \doi{10.1162/tacl_a_00563}.
\newblock URL \url{https://doi.org/10.1162/tacl\_a\_00563}.

\bibitem[Zhang et~al.(2023)Zhang, Press, Merrill, Liu, and Smith]{zhang2023language}
Muru Zhang, Ofir Press, William Merrill, Alisa Liu, and Noah~A Smith.
\newblock How language model hallucinations can snowball.
\newblock \emph{arXiv preprint arXiv:2305.13534}, 2023.

\bibitem[Zhang et~al.(2020)Zhang, Merck, Tsai, Manning, and Langlotz]{zhang2020optimizing}
Yuhao Zhang, Derek Merck, Emily Tsai, Christopher~D Manning, and Curtis Langlotz.
\newblock Optimizing the factual correctness of a summary: A study of summarizing radiology reports.
\newblock In \emph{Proceedings of the 58th Annual Meeting of the Association for Computational Linguistics (ACL)}, 2020.
\newblock URL \url{https://arxiv.org/pdf/1911.02541.pdf}.

\bibitem[Zheng et~al.(2023)Zheng, Dou, Gao, Hua, Shen, Wang, Liu, Jin, Liu, Zhou, Xiong, Chen, Xi, Xu, Lai, Zhu, Chang, Yin, Weng, Cheng, Huang, Sun, Yan, Gui, Zhang, Qiu, and Huang]{zheng2023secrets}
Rui Zheng, Shihan Dou, Songyang Gao, Yuan Hua, Wei Shen, Binghai Wang, Yan Liu, Senjie Jin, Qin Liu, Yuhao Zhou, Limao Xiong, Lu~Chen, Zhiheng Xi, Nuo Xu, Wenbin Lai, Minghao Zhu, Cheng Chang, Zhangyue Yin, Rongxiang Weng, Wensen Cheng, Haoran Huang, Tianxiang Sun, Hang Yan, Tao Gui, Qi~Zhang, Xipeng Qiu, and Xuanjing Huang.
\newblock Secrets of {RLHF} in large language models part {I}: {PPO}, 2023.

\bibitem[Ziegler et~al.(2020)Ziegler, Stiennon, Wu, Brown, Radford, Amodei, Christiano, and Irving]{ziegler2020finetuning}
Daniel~M. Ziegler, Nisan Stiennon, Jeffrey Wu, Tom~B. Brown, Alec Radford, Dario Amodei, Paul Christiano, and Geoffrey Irving.
\newblock Fine-tuning language models from human preferences, 2020.

\end{thebibliography}
\bibliographystyle{iclr2024_conference}

\appendix
\section{Appendix}

\subsection{Prompts}
\label{sec:prompts}
Table~\ref{tab:atomic_q_prompt} contains the prompts used with GPT-3.5 to convert statements into questions for model confidence-based truthfulness estimation.

\begin{table}
    \centering
    \small
    \begin{tabular}{cp{12cm}}
        \toprule
        \makecell{Biography \\ Atomic Fact \\ to Question} & I will provide a statement containing one atomic fact related to Hillary Clinton or people around her. Please rephrase the following statement into a specific question testing knowledge of the key fact in the statement. For example:

Statement: Hillary Clinton was born in 1947.

Question: In what year was Hillary Clinton born?

Statement: Hillary attended the Wellesley College.

Question: What college did Hillary Clinton attend?

Statement: She married Bill Clinton.

Question: Who did Hillary Clinton marry?

I will provide a statement containing one atomic fact related to LeBron James or people around him. Please rephrase the following statement into a specific question that testing knowledge of the key fact in the statement. For example:

Statement: LeBron James is a professional basketball player.

Question: What is LeBron James' profession?

Statement: He is one of the best in the NBA.

Question: Where does LeBron James rank among NBA players?

Statement: James was born in Akron.

Question: In what city was LeBron James born?

I will provide a statement containing one atomic fact related to [NAME] or people around [HIM/HER]. Please rephrase the following statement into a specific question testing knowledge of the key fact in the statement. For example:

Statement: [STATEMENT]

Question: 
\\
        \midrule
        \makecell{MedicalQA \\ Atomic Fact \\ to Question} & I will provide a statement containing one atomic fact about the medical condition menopause. Please rephrase the following statement into a specific question testing knowledge of the key fact in the statement. For example:

Statement: Menopause is a time in a woman's life.

Question: Menopause is a time in whose life?

Statement: Menopause is the time when a woman no longer has menstrual periods.

Question: Menopause is the time when a woman no longer has what?

Statement: There is a decline in the ovarian hormone estrogen.

Question: During menopause there is a decline in what?

I will provide a statement containing one atomic fact about the medical condition breast cancer. Please rephrase the following statement into a specific question testing knowledge of the key fact in the statement. For example:

Statement: The signs and symptoms include a lump or thickening in or near the breast or underarm.

Question: Do the signs and symptoms of breast cancer include a lump or thickening in or near the breast or underarm?

Statement: The signs and symptoms include a change in the size or shape of the breast.

Question: Do the signs and symptoms of breast cancer include a change in the size or shape of the breast?

I will provide a statement containing one atomic fact about the medical condition varicose veins. Please rephrase the following statement into a specific question testing knowledge of the key fact in the statement. For example:

Statement: Varicose veins occur when the veins under the skin become enlarged.

Question: Varicose veins occur when what happens to the veins under the skin?

Statement: Veins in the legs lose their ability to efficiently circulate blood back to the heart.

Question: Varicose veins in the legs lose their ability to efficiently do what?

I will provide a statement containing one atomic fact about the medical condition [CONDITION]. Please rephrase the following statement into a specific question testing knowledge of the key fact in the statement. For example:

Statement: [STATEMENT]

Question:
 \\
        \bottomrule
    \end{tabular}
    \caption{GPT-3.5 prompts used to convert atomic facts into questions.}
    \label{tab:atomic_q_prompt}
\end{table}

\subsection{Sample Model Generations}
See Tables~\ref{tab:examples-greta} and~\ref{tab:examples} for samples generated by several different models. After factuality tuning, the model does produce somewhat terser responses.

\begin{table}
    \centering
    \small
    \begin{tabular}{cp{12cm}}
        \toprule
        SFT & Greta Gerwig is an American actress and filmmaker. She was born in and raised in Sacramento, California. Gerwig became interested in the film industry in 2002, when she attended the San Francisco Art Institute. She got her start in 2006, when she starred in the film `The Off Hours'. Her other films include Frances Ha, Mistress America, and 20th Century Women. In 2018, Gerwig directed Lady Bird, and received wide acclaim, a Golden Globe and Tony nomination for directing, and Oscar and BAFTA nominations for the film. \\
        \midrule
        FactTune-FS & Greta Gerwig is an American actress and filmmaker. Born in Sacramento, California in 1983, Gerwig started acting at a young age. She is best known for starring in films such as `Frances Ha' and `Mistress America.' Gerwig was nominated for an Oscar for Best Actress for her performance in `Lady Bird.' Gerwig has written and directed a number of acclaimed independent films, such as `Frances Ha' and `Mistress America.' \\ \midrule 
        FactTune-MC & Greta Gerwig is an American actress and filmmaker. She was born in 1983 in Sacramento, California, and raised in Stanford, California and New York City. Gerwig has appeared in such movies as Greenberg, Frances Ha, and Jackie. She also portrayed the title role in the 2013 film of Jackie Kennedy. Her directorial debut is the 2012 movie, Lady Bird, which was nominated for the Academy Award for Best Picture. \\
        \bottomrule
    \end{tabular}
    \caption{Samples for the prompt ``Write a biography of Greta Gerwig" under the SFT, FactTune-FS, FactTune-MC Llama-1-7b models.}
    \label{tab:examples-greta}
\end{table}

\begin{table}
    \centering
    \small
    \begin{tabular}{cp{12cm}}
        \toprule
        SFT & Mukesh Ambani was born in 1957, in Mumbai, India, to billionaire industrialist Dhirubhai Ambani and Kokilaben Ambani. Mukesh Ambani attended Stanford University, where he received a Bachelor of Science degree in Chemical Engineering. He is the chairman and managing director of Reliance Industries Limited (RIL), a company owned by the Ambani family of India. In 2002, Ambani became the chairman of Reliance Natural Resources Limited, a company producing oil and gas. Reliance has been one of the largest retailers in India since 2006, with 380 Reliance Fresh outlets and 1,250 stores under Reliance Retail. Reliance is the world's largest polyester fiber manufacturer, operating 98 integrated manufacturing sites worldwide.\\
        \midrule
        FactTune-FS & Mukesh Ambani is an Indian business magnate and the chairman, managing director and largest shareholder of Reliance Industries Limited (RIL), a Fortune Global 500 company and India's most valuable company by market value. Ambani is the richest man in India, with a net worth of US\$27.4 billion, according to Forbes. Mukesh Ambani was born on April 19, 1957 in Aden, Yemen. He is the eldest son of Indian business magnate Dhirubhai Ambani. Mukesh Ambani currently lives in Mumbai, India. \\ \midrule 
        FactTune-MC & Mukesh Ambani is an Indian businessman and the CEO of Reliance Industries. He is the richest person in India and the 19th richest person in the world. Ambani was born in 1957 in Aden, Yemen. He moved to India with his family in 1958. Ambani studied chemical engineering in the U.S and worked in the oil company, Reliance, which is the largest Indian company. \\
        \bottomrule
    \end{tabular}
    \caption{Samples for the prompt ``Write a biography of Mukesh Ambani" under the SFT, FactTune-FS, FactTune-MC Llama-1-7b models.}
    \label{tab:examples}
\end{table}

\end{document}